\documentclass{article}
\usepackage{ijcai19}

\usepackage{times}
\usepackage{soul}
\usepackage{url}
\usepackage[hidelinks]{hyperref}
\usepackage[utf8]{inputenc}
\usepackage[small]{caption}
\usepackage{graphicx}
\usepackage{amsmath}
\usepackage{booktabs}
\usepackage{algorithm}
\usepackage{algorithmic}
\title{Time Series Anomaly Detection Using Convolutional Neural Networks and Transfer Learning}

\author{
Tailai Wen$^1$
\and
Roy Keyes$^2$
\affiliations
Arundo Analytics
\emails
\{$^1$tailai.wen, $^2$roy.keyes\}@arundo.com
}

\begin{document}

\maketitle

\begin{abstract}
  Time series anomaly detection plays a critical role in automated monitoring systems. Most previous deep learning efforts related to time series anomaly detection were based on recurrent neural networks (RNN). In this paper, we propose a time series segmentation approach based on convolutional neural networks (CNN) for anomaly detection. Moreover, we propose a transfer learning framework that pretrains a model on a large-scale synthetic univariate time series data set and then fine-tunes its weights on small-scale, univariate or multivariate data sets with previously unseen classes of anomalies. For the multivariate case we introduce a novel network architecture. The approach was tested on multiple synthetic and real data sets successfully.
\end{abstract}

\section{Introduction}
Time series anomaly detection plays a critical role in automated monitoring systems. It is an increasingly important topic today, because of its wider application in the context of the Internet of Things (IoT), especially in industrial environments~\cite{da2014internet}.

Before the boom of deep learning in the early 2010s, most time series anomaly detection efforts were based on traditional time series analysis (e.g.~\cite{abraham1989outlier,bianco2001outlier}), or on approaches to extract and represent time series properties (e.g.~\cite{chan2005modeling,keogh2005hot,ringberg2007sensitivity,ahmed2007multivariate}). Some machine learning techniques for multivariate outlier detection were also widely applied to detect anomalies in multivariate time series (e.g.~\cite{breunig2000lof,scholkopf2001estimating,liu2008isolation}), although they treat data points independently and neglect temporal relationships.

In recent years, advances in deep learning have revolutionalized many areas of data-driven modeling. Recurrent neural networks (RNN) and convolutional neural networks (CNN) are two major types of network architectures that enabled these breakthroughs. RNN's are generally applied to temporal sequence tasks, while CNN's are typically the first choice for image related tasks. For this reason, most previous deep learning work on time series anomaly detection was based on RNN's (e.g.~\cite{malhotra2015long,kim2016long,wang2017time,yin2017deep}). There has been some research using CNN's for time series tasks, primarily around sequence classification (e.g.~\cite{zheng2014time,yang2015deep,cui2016multi,rajpurkar2017cardiologist}).

We recognized that time series anomaly detection shares many common aspects with image segmentation. When a person visualizes a time series and selects the anomalous segment, if present, the perceptual process is very similar to a person looking at an image and marking a desired object. In this research, we created a CNN-based deep network for time series anomaly detection. In particular, we were inspired by a successful image segmentation network, U-Net, and applied a time series version of U-Net to detect anomalous segments in time series. As the limited occurrence of failures is a common blocker for anomaly detection in industrial IoT systems, we also propose a transfer learning framework to resolve the data sparsity issue, including a new architecture, MU-Net, for transferring a univariate base model to multivariate tasks.

The remainder of this paper is organized as follows: Section~\ref{relatedwork} includes a review of related work. We provide details of applying U-Net for time series anomaly detection in Section~\ref{unetsection}. Section~\ref{transfer} introduces the transfer learning framework and introduces the MU-Net architecture for multivariate time series anomaly detection. We provide experimental observations in Section~\ref{test}, and conclude with some discussion in Section~\ref{conclusion}.

\begin{figure*}[hbt!]
  \centering
  \includegraphics[scale=0.35]{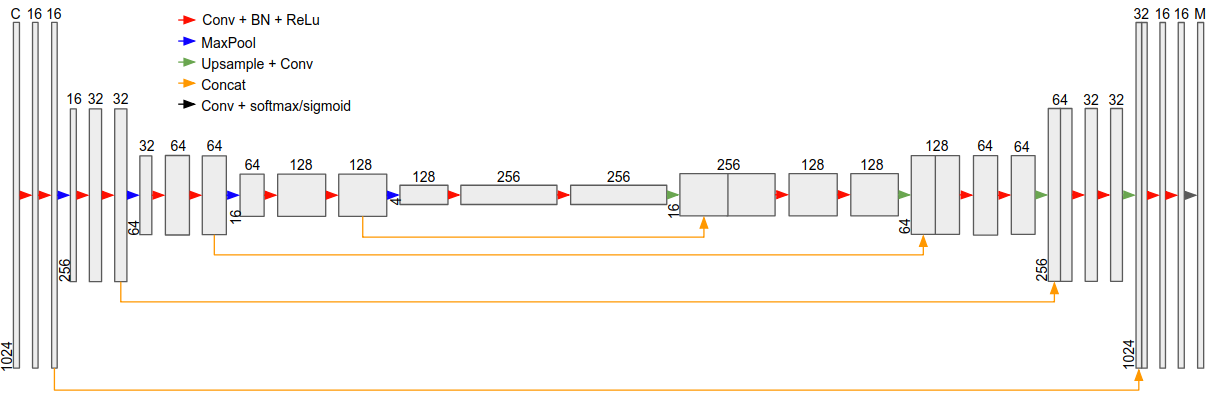}
  \caption{Architecture of U-Net for time series segmentation.}
  \label{unet}
\end{figure*}
\section{Related Work}\label{relatedwork}
Our approach was influenced by recent successes of deep learning for image segmentation. \cite{long2015fully} proposed a fully convolutional network (FCN), where common convolutional architectures for image classification (e.g. AlexNet, the VGG net, and GoogLeNet) are used as encoders, and counterpart deconvolution layers are used for upsampling as decoders. U-Net \cite{ronneberger2015u} improved upon the FCN architecture by introducing so-called skip channels between encoding layers and decoding layers into the architecture, so that high-level features and low-level features are concatenated to prevent information loss along deep sequential layers. This architecture was proven successful when applied to segmentation of neuronal structures in electron microscopic images in the original paper. It was subsequently applied to several other biomedical image segmentation tasks (e.g.~\cite{cciccek20163d,dalmics2017using,dong2017automatic}) as well as image segmentation problems in earth science (e.g.~\cite{karchevskiy2018automatic}), remote sensing (e.g.~\cite{yao2018pixel}), and automated driving (e.g.~\cite{siam2017deep}).

Our transfer learning work follows the success of deep transfer learning in image and natural language processing. The design of synthetic pre-training data was in part inspired by \cite{mahajan2018exploring}. Our fine-tuning strategies incorporated some techniques presented in the Fast.AI courses\footnote{\url{https://www.fast.ai}}.

Data augmentation proved to be an important part of the research. Our time series augmentation strategies extended the previous work of \cite{um2017data}.

\section{CNN-based Time Series Segmentation}\label{unetsection}
We will introduce the details of how we build and train a time series version of U-Net, including the detailed architecture, how it is applied to streaming data in a production environment, and some important issues, including input normalization and augmentation.

\subsection{U-Net for time series segmentation}
A time series can be regarded as a one-dimensional image where the only dimension is temporal, whereas a typical image has two dimensions: width and height. A multivariate time series may have an arbitrary number of channels, which may have different properties and correlations with each other. In contrast, an image typically has only three channels, RGB, and their properties and correlations are not arbitrary. Following the design of U-Net, we propose the following architecture for time series segmentation as shown in Figure~\ref{unet}. For a time series with length 1024 and $C$ channels, it is encoded by five sections of convolution layers. Each section includes two layer blocks, each of which includes a convolution layer, a batch normalization layer, and a ReLu layer. For all convolution layers, the kernel size is equal to 3. The number of filters in each convolution layer increases over the five sections, from 16 to 256 by a factor of 2. For every convolution layer, we apply zero padding and set the stride equal to 1. Between encoding sections, max pooling layers with a pool size equal to 4 are used to downsample the feature series. This is followed by four decoding sections, where each section includes an upsampling layer and two conv+BN+ReLu blocks. The upsampling rate is also equal to 4, and the convolution layers uses the same number of filters as their counterpart encoding layers, making the architecture symmetric. An important feature of U-Net is the skip channels between corresponding encoding and decoding sections. This is implemented by concatenating the output from every max pooling layer with the output from the corresponding upsampling layer, and performing convolution over the concatenated feature series.

The last section includes a convolution layer with a kernel size equal to 1 and an activation layer. Sometimes different classes of anomalies are not mutually exclusive, and therefore the problem is a multi-class, multi-label problem, i.e. a time point can be assigned to multiple classes. For example, a spike on a periodic series is both an additive anomaly and a seasonal anomaly. In such a case, the number of filters at the final convolution layer is equal to the number of anomaly classes $M$, and the final activation function is a sigmoid so that probabilities of classes are independent. If all classes of anomaly are mutually exclusive, the output shape has depth $M+1$ and the additional column is for the nominal class (i.e. no anomaly present). In this case, softmax activation is applied so that the result is multi-class single-label. In both cases, soft Dice loss is used as the loss function, as is the case with many image segmentation networks, and the Adam optimizer is used.

\subsection{Prediction on Streaming Data}
\begin{figure}[hbt!]
  \centering
  \includegraphics[scale=0.45]{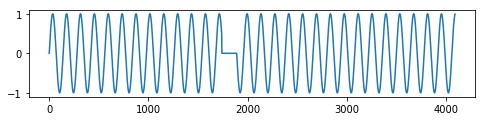}
  \includegraphics[scale=0.45]{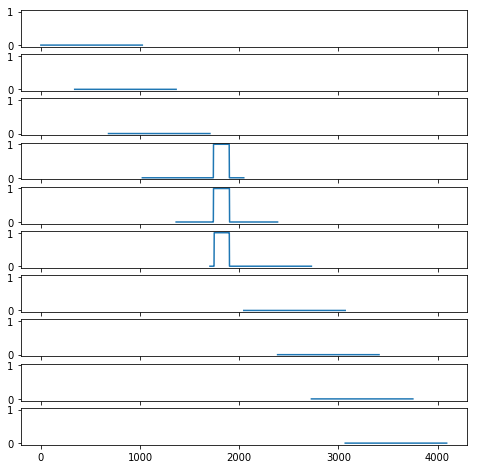}
  \caption{Anomaly detection on a data stream (top) by taking snapshots regularly and returning probability of anomaly (bottom).}
  \label{slide}
\end{figure}

When deploying a trained model for production in a streaming environment, we regularly take snapshots of the latest batch of data and run the model on it. The frequency of taking snapshots should be at least as high as the length of snapshot so that every time point is evaluated by the model at least once. We recommend using a frequency several times higher than this minimal limit, and thus every time point is evaluated by the model a few times and we may ensemble multiple results for the same time point to get more robust detection. Figure~\ref{slide} shows an example where every data point is evaluated 3 times.

Although the model input shape is fixed (e.g. 1024), it is not necessary to always use it as the length of the snapshot. The length of the snapshot should be determined by the streaming frequency and the time scale of anomalous behavior. If the streaming frequency is too high with respect to anomalous behaviors, a long snapshot length (i.e. a large number of time points) should be used and every snapshot needs to be downsampled to the model input size. If the streaming frequency is too low, a short length should be used and upsampling is needed.

The strategy of sliding window could also work with a CNN-based classification model that classifies sequences by whether it includes anomalous subsequence. However, the choice of snapshot length would be problematic. If it is too long, the localization of anomaly segments tends weak, since a snapshot is only classified by not segmented. On the other hand, if it is too short, the snapshot may not contain sufficient context to distinguish itself from normal subsequence. The proposed segmentation-based method, however, may overcome this difficulty by segmenting anomalous period in high granularity.

\subsection{Input Normalization}
\begin{figure}[hbt!]
  \centering
  \includegraphics[scale=0.5]{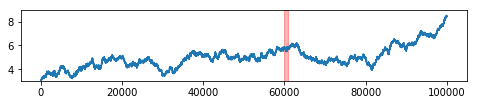}
  \includegraphics[scale=0.5]{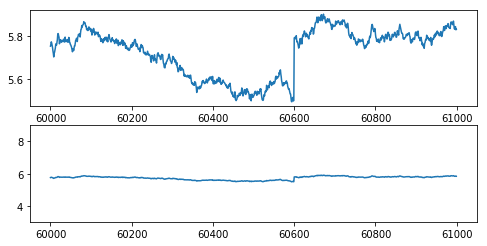}
  \caption{The same segment from a time series (top) in different scales (middle and bottom).}
  \label{input_normal}
\end{figure}

Different from usual image problems, where values in a RGB channel are always between 0 and 255, a time series may have arbitrarily large or small values, and the magnitude scale may vary over channels. Figure~\ref{input_normal} shows the same subsequence from a stock price series in two different scales, where one looks essentially constant, while the other appears to have a big jump. This is a common misleading issue, even when a human views a time series. The proposed model requires a user to specify a magnitude scale first, and uses the scale to normalize the input series. Therefore, when applying the model in production over snapshots, the detection will always be based on the same scale. In some cases, the user does need to detect anomalies only with respect to values inside a snapshot. The user may then add a sample-wise input normalization layer at the start of the neural network, so that every input snapshot is normalized independently and the scale will be dynamic.

\subsection{Augmentation}\label{aug}
Data augmentation is a common approach to boost the size of training data for robust models. \cite{um2017data} introduced a list of augmentation methods for time series, including time warping, cropping, etc. We extended the list by adding a few methods, for example zooming, adding random trend, reversing series, applying a random linear operation, random mutation between multiple series, etc. These augmentation methods were used during testing over different data sets.

As in image problems, augmentation of training data must be label-invariant. In the context of anomaly detection, that means an augmentation method must not change the nominality of a time series. Some augmentation operations are not label-invariant to certain types of anomaly. For example, mutation operation may violate a periodic pattern when applied to a seasonal series, so it should not be used to augment training data if the expected anomaly is an aberration of a periodic pattern. Augmentation methods should always be selected appropriately for the case under consideration.

\section{Transfer Learning}\label{transfer}
Sparsity of failure events in historical data often limits model training in practice. Transfer learning is a strategy to resolve the data sparsity issue. The transfer learning approach uses the weights from a base model pretrained on an available large-scale data set and then fine-tunes the model weights with a small-scale data set related to the task of interest. The hyper-parameter tuning process takes advantage of the model pretrained on the large-scale data set, which tends to extract useful features from the input, and therefore requires dramatically less training data to converge without overfitting.

\cite{mahajan2018exploring} explored factors that may impact performance of transfer learning in CNN-based image processing tasks. The similarity between the pretraining data set and target data set has proven to be an important factor. The correlation between pretraining task(s) and the target task also plays a significant role. In our work, we define three pretraining tasks, i.e. three types of anomalies to detect, including additive outliers, anomalous temporary changes of volatility, and violations of cyclic patterns. These are the most common types of anomalies found in univariate time series. Additive outliers can be interpreted on a short time scale, violations of cyclic patterns must be interpreted on a long time scale that covers at least a few cycles, and changes of volatility occurs on a medium time scale. We believe low-order features representing these anomaly types are necessary components to build higher-order features to represent more complex anomalous behaviors. In other words, those features are transferable to general the anomaly detection problem.

As features are not only extracted from anomalous segments, but also nominal segments, and thus, the nominal behavior of pretraining series should also be diverse. To prepare the pretraining data set, we generated a large number of various synthetic time series, including smooth curves, piece-wise linear curves, piece-wise constant curves, pulse-like signals, etc. For each type, we generated some cyclic series and some non-cyclic series. We augmented those time series by cropping, jittering, adding random trends, and time warping, with different levels of intensity, such that the training data covers a variety of nominal behaviors. We then adjusted random segments in these nominal series with the three previously mentioned types of anomalies to generate pretraining data with labelled anomalies.

\begin{figure}[hbt!]
  \centering
  \includegraphics[scale=0.45]{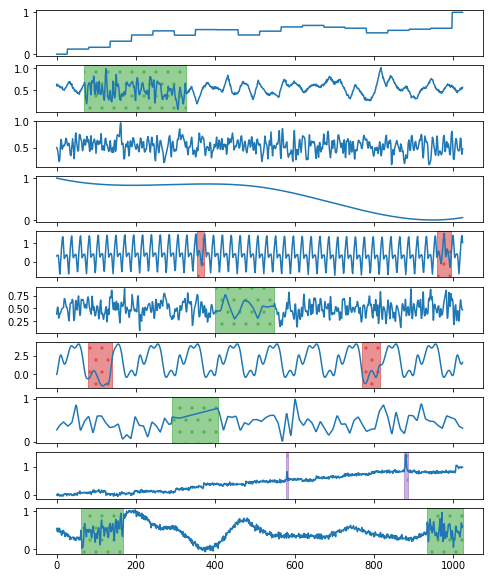}
  \caption{Some samples from the pretraining data set with labelled additive outliers (purple), changes of volatility (green), and violations of cyclic patterns (red).}
  \label{pretraining}
\end{figure}

Figure~\ref{pretraining} shows some examples from the pretraining data set.

\subsection{Transfer Learning for Univariate Tasks}
For transfer learning to another univariate task, we keep the model architecture as in Figure~\ref{unet}, except that the output shape of the final convolution layer must change according to the number of classes in the target task. Weights of all layers except the output section are initialized with weights from the pretrained model. We found two fine-tuning strategies with good performance in our tests. The first one is to set up different learning rate multipliers in 10 sections (5 encoding sections, 4 decoding sections, and the output section) as 0.01, 0.04, 0.09, ..., 0.81, 1.0. The other one is to freeze the weights in the first two sections and only fine-tune the subsequent sections, and then to unfreeze the first sections and fine-tune all weights. Both strategies returned similar results in our tests.

\subsection{MU-Net: A U-Net-based Network for Transfer Learning from Univariate to Multivariate Tasks}
When the pretrained base model is transferred to a multivariate task, using the same U-Net architecture would be problematic. The kernel of the first convolution layer would have a different shape, as $C$ is not equal to 1. If those kernels are initialized randomly, then transferring weights of the deeper layers is meaningless, because the lowest-order features are extracted differently. If we repeat the $3\times 1 \times 16$ weight matrix in the first layer of pretrained model $C$ times and create an initialization of the $3\times C \times16$ weight matrix for the transfer model, mathematically it is equivalent to extracting transferable features from the sum of all series channels. While the sum of RGB channels may still maintain important properties from original images, the sum of time series channels is generally meaningless, for example, if the channels represent different sensors in an IoT system such as pressure and temperature.
\begin{figure*}[hbt!]
  \centering
  \includegraphics[scale=0.35]{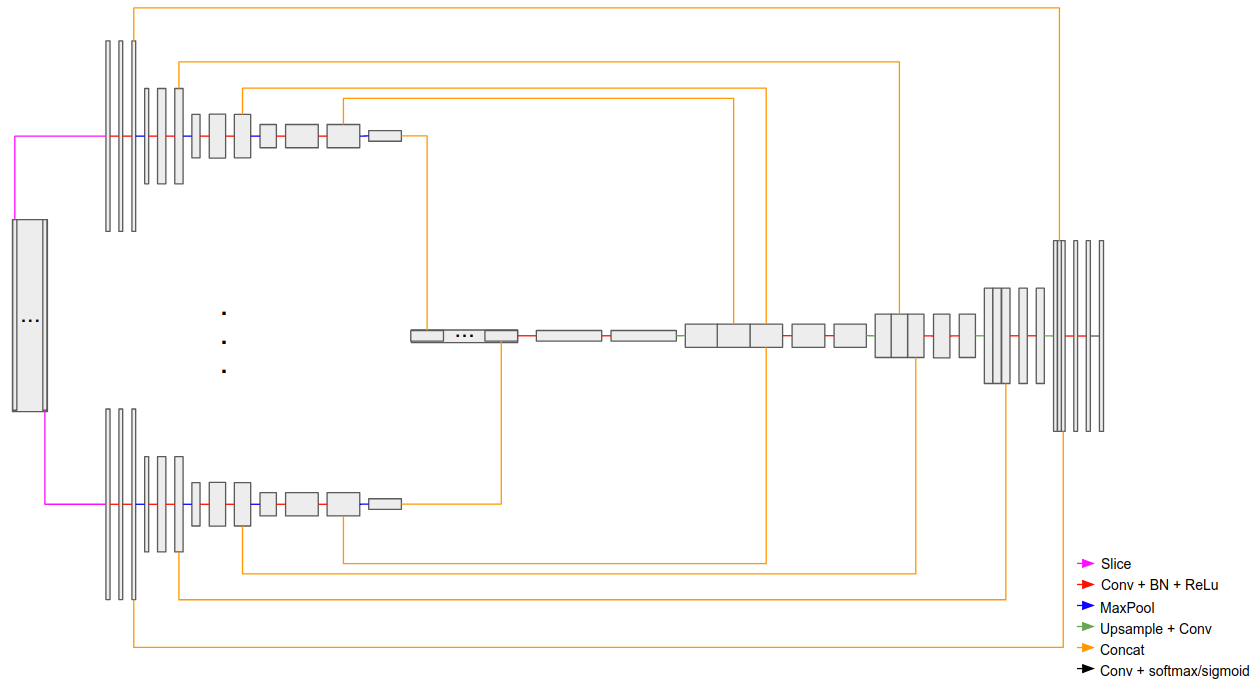}
  \caption{Architecture of MU-Net, a U-Net-based transfer network for multivariate tasks.}
  \label{onet}
\end{figure*}

We propose a new network architecture to transfer the weights from a pretrained model as shown in Figure~\ref{onet}. Similar to U-Net, this architecture also has an encoding-decoding structure. However, the $C$ channels of input series are separated by a slicing layer first, and then every channel has its own univariate encoding sections, like the first four encoding sections in U-Net. The outputs from the fourth encoding section over all channels will be concatenated before an integrated fifth encoding section, followed by four decoding sections and a final output section, the same as U-Net. We have nicknamed this architecture MU-Net (multivariate U-Net).

When we transfer a pretrained U-Net model to a multivariate task, we initialize weights in the first four encoding sections for every channel in MU-Net with the corresponding weights in the pretrained model. During fine-tuning, we first freeze the first four encoding sections and tune the weights of the remaining layers. We then unfreeze the third and fourth encoding sections and continue tuning. Finally we tune all weights including the first two sections.

\section{Experimental Evaluation}\label{test}
We tested the proposed approach in four scenarios: a univariate task with sufficient data, a multivariate task with sufficient data, a univariate task with insufficient data and transfer learning, and a multivariate task with insufficient data and transfer learning.

All tests were conducted on an Nvidia GeForce GTX 1080Ti GPU. All implementation was done using Keras with the TensorFlow backend.

\subsection{Dodgers Loop Sensor Data Set}
This data set\footnote{\url{https://archive.ics.uci.edu/ml/datasets/Dodgers+Loop+Sensor}} was originally introduced by~\cite{ihler2006adaptive}. It includes a single time series of the traffic on a ramp close to Dodger Stadium over 28 weeks with 5-min frequency. The task is to detect anomalous traffic caused by sporting events. There are a total of 81 known events. We use the first half of the data (including 42 events) to train the model and the other half (including 39 events) for testing.

\begin{figure}[hbt!]
  \centering
  \includegraphics[scale=0.45]{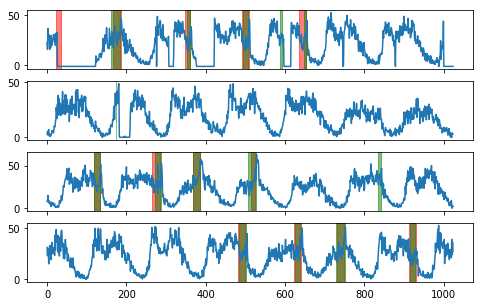}
  \caption{Some snapshots in the Dodgers test case. Red markers represent known events, green markers represent model detection.}
  \label{dodger}
\end{figure}

The training sequence was randomly cropped into 500 snapshots with lengths varying between 1,024 (about 3.5 days) and 4,096 (about 2 weeks), and then downsampled to 1,024. A univariate U-Net was created and trained. Only 3 known events out of 39 were not detected, and the missing detections were because these events were very close to periods with many missing values. A few false positives also occurred, mostly near missing values. Figure~\ref{dodger} shows detection results in some snapshots from the testing set.

\subsection{Gasoil Plant Heating Loop Data Set}
This data set\footnote{\url{https://kas.pr/ics-research/dataset_ghl_1}} was originally introduced by~\cite{filonov2016multivariate}. It includes 48 simulated control sequences of a gasoil plant heating loop, which suffered cyber-attacks at some points. All time series have 19 variables and an average length of 204,615. We used the tag DANGER as the indicator of attack events.

We used 30 sequences for model training and the remaining 18 for testing. The training data was randomly cropped into 300 snapshots with length 50,000, and then downsampled to 1,024. A multivariate U-Net ($C=19$) was created and trained. Among the 18 testing sequences with 22 cyber-attacks in total (14 sequences with 1 attack, 4 with 2 attacks), only 1 attack was missed by detection. There were 3 false alarms.

\subsection{Synthetic Curves with Unusual Shapes}
\begin{figure}[hbt!]
  \centering
  \includegraphics[scale=0.45]{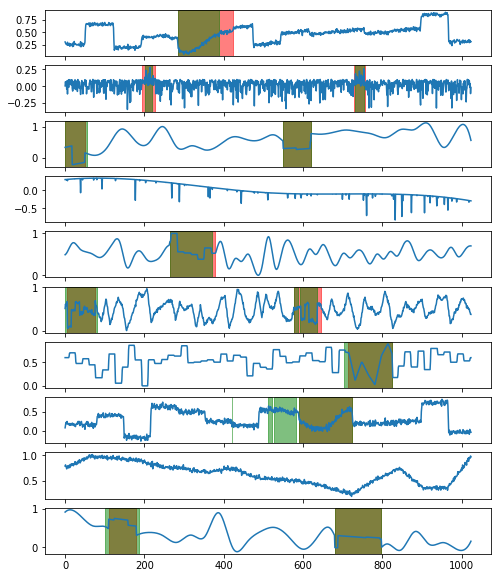}
  \caption{Some examples of synthetic curves with unusual shapes. Red markers represent true anomalies, green markers represent model detection.}
  \label{shape}
\end{figure}

We created a data set of synthetic curves where each series may have one or several segments with different curve shape from the rest of that series. For example, a series could be mostly smooth except for a segment that is piece-wise linear. We augmented the data set with several augmentation methods mentioned in Section~\ref{aug}, so that series behaviors are diverse. We generated 1,400 samples with length 1,024: 700 for training and 700 for testing. Figure~\ref{shape} shows some examples. The task is to detect unusual segments, which is a more challenging task than the pretraining tasks. Although 700 is larger than the size of data sets in the previous two testing cases, it is still insufficient considering the diversity of time series behaviors over these samples.

We used intersection over union (IoU) score to evaluate the performance of this task. When training a U-Net from scratch, the testing IoU was 50.96\%. When we used univariate transfer learning, the testing IoU reached 71.95\%. Some results from the transferred model are shown in Figure~\ref{shape}.

\subsection{Electromyography (EMG) Data Set}
This data set\footnote{\url{http://archive.ics.uci.edu/ml/datasets/EMG+data+for+gestures}} was originally introduced by~\cite{lobov2018latent}. It includes 8-channel myographic signals recorded by bracelets worn on 36 subjects' forearms. Each subject performed the same set of seven gestures twice sequentially. The task is to detect different gestures. Precisely speaking, this task is not an anomaly detection task but a segmentation one, because negative labels (i.e. segments with no gesture) are much less frequent than positive labels (i.e. segments with a gesture). However, since the approach we propose is fundamentally a time series segmentation method, we believe this is still a good case for testing. This case is the only testing case assuming a multi-class, single-label scenario, because we know there was at most one gesture performed at a time point.

\begin{figure}[hbt!]
  \centering
  \includegraphics[scale=0.45]{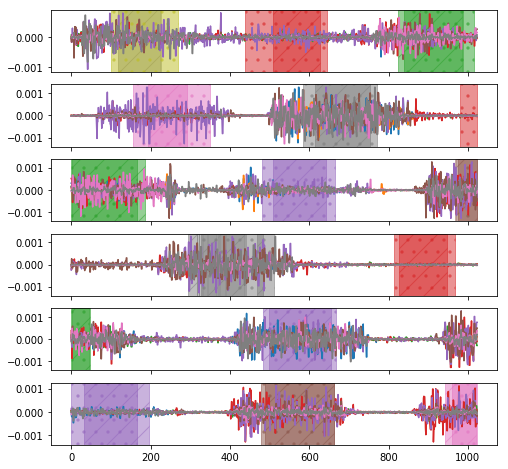}
  \caption{Some snapshots in the EMG test case. Dotted shadows represent true segments, and striped shadows represent model segments. Different colors represent different gestures.}
  \label{emg}
\end{figure}

We trained a U-Net from scratch, and the IoU score evaluated on the test set was 56.61\%. We also trained an MU-Net from scratch, and the IoU score was 64.10\%. We then transferred the pretrained U-Net to a MU-Net and fine-tuned the weights, the IoU score then reached 70.04\%. Figure~\ref{emg} shows some testing snapshots with results from the transferred model.

\section{Conclusion and Discussion}\label{conclusion}
In this work, we proposed a time series version of the convolutional U-Net for time series anomaly detection. As far as we are aware of, this is the first work to use a CNN-based deep network for time series segmentation in the context of anomaly detection. The architecture was tested with both univariate and multivariate examples and showed satisfactory performance.

To address the challenge of data sparsity that often occurs in real-world anomaly detection tasks, we proposed a transfer learning framework to transfer a U-Net model pretrained on a large-scale univariate time series set to general anomaly detection tasks. In particular, to transfer to a multivariate task, we proposed a new architecture, MU-Net, that may take advantage of the pretrained univariate U-Net. The transfer learning framework was also tested with both univariate and multivariate examples and returned promising results.

One of the primary, inherent challenges in time series anomaly detection is defining ground truth. For time series, delineating exactly when anomalous behavior occurs and when it stops is a fundamental difficulty, as even human experts are likely to differ in their assessments. Additionally, when detecting anomalies in time series, there is the question of what counts as a useful detection. For a typical image segmentation task, the goal is to reproduce segmentation masks as created by human labellers. For image tasks the level of ambiguity is relatively low, i.e. edges are relatively well defined. For time series anomaly detection, the predicted mask and the ground truth mask may have less than perfect overlap, but, operationally, the predicted mask may still serve its purpose of alerting to the presence of a specific type of anomaly. Ultimately this means that the quantifying the goodness of time series anomaly detection via segmentation is difficult.

A transferable model is particularly well-suited for handling the large variety of possible behaviors in time series data. The pretraining dataset was populated with what we considered to be the most fundamental anomalous behaviors in time series data. Because the definition of anomalous behavior is context-dependent, the pretraining dataset was designed to train a model capable of extracting informative, underlying features for various types of time series and anomalies, such that a new anomaly type may be easily learned from a small-scale dataset. This methodology is fundamentally different from most previous work on anomaly detection, which were based on identifying outliers with an explicit or implicit definition. While this scenario is less suited to handling previously unseen anomalies than more traditional outlier detection based methods, its strength lies in its robustness and ability to handle complex signals.

Future work on this topic could include extensive experiments on factors that may impact transfer learning performance, such as \cite{mahajan2018exploring} performed for image classification. That paper explored, given specific target tasks, how a pretrained model may perform differently if it was pretrained on ImageNet, or a certain part of ImageNet, or an even larger and more comprehensive data set. The correlation between different types of time series (on both nominal and anomalous behaviors) is even more subtle than that between image classes. We believe that more research in this direction would help us improve the pretraining data set as well as the transfer learning framework, such that the pretrained model would be even more transferable.

Another area of possible future work is to create performance comparisons with benchmark algorithms, such as statistical time series analysis, RNN-based anomaly detection methods, and CNN-based classification methods with sliding windows.

\section*{Acknowledgments}
We would like to thank Jason Hu and Gunny Liu for their help with synthetic data generation. We also thank Fausto Morales, Pushkar Kumar Jain, and Henry Lin for helpful discussions.

\bibliographystyle{named}
\bibliography{ijcai19}

\end{document}